\documentclass[letterpaper, 10 pt, conference]{ieeeconf}  

\IEEEoverridecommandlockouts                              
\usepackage{graphics}
\usepackage{amsmath}
\usepackage{graphicx}
\usepackage{float}
\usepackage[caption = false]{subfig}
\usepackage{longtable}
\usepackage{booktabs}
\usepackage{multirow}
\usepackage{siunitx}
\usepackage{algorithm}
\usepackage[noend]{algpseudocode}
\usepackage{mathpazo}

\makeatletter
\def\BState{\State\hskip-\ALG@thistlm}
\makeatother

\title{\LARGE \bf
Siamese networks for generating adversarial examples  
}

\author{ \parbox{3 in}{\centering Mandar Kulkarni\\
        Data Scientist\\
        Schlumberger\\
        {\tt\small mkulkarni9@slb.com}}
        \hspace*{ 0.5 in}
        \parbox{3 in}{ \centering Aria Abubakar\\
        Data Analytics Program Manager\\
        Schlumberger\\
        {\tt\small aabubakar@slb.com}}
}

\begin{document}

\maketitle
\thispagestyle{empty}
\pagestyle{empty}

\begin{abstract}
Machine learning models are vulnerable to adversarial examples. An adversary 
modifies the input data such that humans still assign the same label, however,   machine learning models misclassify it. Previous approaches in the literature demonstrated that adversarial examples can even be generated for the remotely hosted model. 
In this paper, we propose a Siamese network based approach to generate adversarial examples for a multiclass target CNN. We assume that the adversary do not possess any knowledge of the target data distribution, and we use an unlabeled mismatched dataset to query the target, e.g., for the ResNet-50 target, we use the Food-101 dataset as the query. Initially, the target model assigns labels to the query dataset, and a Siamese network is trained on the image pairs derived from these multiclass labels. We learn the \emph{adversarial perturbations} for the Siamese model and show that these perturbations are also adversarial w.r.t. the target model. In experimental results, we demonstrate effectiveness of our approach on MNIST, CIFAR-10 and ImageNet targets with TinyImageNet/Food-101 query datasets.


\end{abstract}

\section{Introduction and related works}
Machine learning models are vulnerable to external attacks such as adversarial inputs. 
The examples from the dataset can be perturbed in a manner that a human assigns the same label to it, however, it machine learning models to misclassify it. Such examples are referred to as the Adversarial examples. 
Szegedy et al. \cite{szegedy2013intriguing} proposed a box-constraint optimization to obtain the adversarial samples from the input images. In addition, it was observed that the same perturbation causes different networks trained on different subsets of the dataset to
misclassify the same input. Thus, adversarial examples are transferable between the models trained on similar data distribution.
Goodfellow et. al. \cite{goodfellow2014explaining} proposed fast gradient sign (FGS) method for generating adversarial examples. Small input perturbations that maximize the loss, locally, produce images that are more often misclassified by the network. It was argued that the primary cause of the vulnerability of neural networks to adversarial perturbation is their linear nature.
Moosavi et al. \cite{moosavi2017universal} discovered the existence of a universal (image-agnostic) perturbation that causes natural images
to be misclassified with high probability. The existence of universal perturbations point towards the
geometric correlations among the high-dimensional
decision boundary of classifiers. It underlines the existence of single directions
in the input space that adversaries can possibly exploit to enforce a classifier to make mistakes on natural images. 

\begin{figure}
\includegraphics[width=\columnwidth]{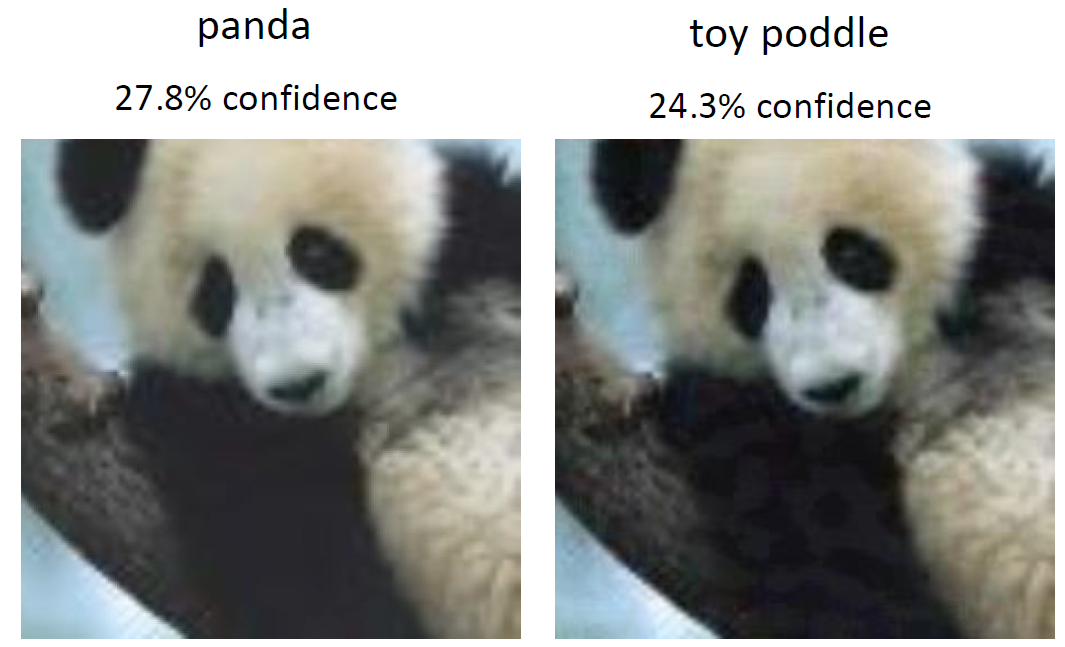}
\caption{\label{fig:inpt117} Example of an adversarial image generated by our approach for RESNET-50 model.}
\end{figure}


Adversarial examples pose a potential threat to machine learning models when deployed in the real world. Since these perturbations tend to go undetected by the humans, it becomes difficult to rectify them. Kurakin et al. \cite{kurakin2016adversarial} showed that even in physical-world scenarios, machine learning systems are vulnerable to adversarial examples. It was demonstrated that a large fraction of the adversarial examples obtained from a cell-phone camera get misclassified by the ImageNet Inception model. 



It was believed that the adversarial samples can only be generated when we have an access to the target model. However, Papernot et al. \cite{papernot2017practical} demonstrated that the adversarial examples can even be generated for the remotely hosted deep neural network (DNN) with no knowledge of the architectural of the DNN, such as the number, type, and size of layers, nor of the training data used to learn the DNN's parameters.
The only access their adversary has to the target model is the (atomic) class label of the submitted query vector. They refer to this setting as the black-box scenario.
Starting with a small subset of a training set, a Jacobian-based dataset augmentation technique is used to effectively learn the decision boundary with the limited number of queries made to the target DNN.  

In this paper, we propose an approach based on Siamese networks \cite{koch2015siamese}\cite{chopra2005learning} to generate adversarial examples for the target classification model. Instead of learning the decision boundary of the target model, we directly learn the input perturbation that changes the label of the images.  
We use Siamese networks to learn such adversarial perturbation and show its effectiveness in generating adversarial examples for the target model. 
We use an unlabeled mismatched image dataset as query to the target convolutional neural network (CNN). 
We refer to the mismatched dataset as the "stimulus". Kulkarni et al. \cite{kulkarni2017knowledge} showed the effectiveness of mismatched stimulus for knowledge distillation. We input the stimulus dataset to the target model and obtain the corresponding class labels. Based on the similarity and dissimilarity of the assigned labels, we generate positive and negative image pairs for training the Siamese network. We learn the \emph{adversarial perturbations} for the Siamese network using the FSG method \cite{goodfellow2014explaining} in which  perturbations are applied only to the left image channel of the network to 'switch' the original label. We apply the learned perturbations to the target images and show that they act as adversarial perturbations. 

Our experimental setting is similar to \cite{papernot2017practical} with a few differences. Instead of remote DNN APIs, we demonstrate results with pretrained CNN  models. Also, we do not assume availability of the subset of original training set. Through the use of the Siamese network, we utilize the notion that the target CNN assigns different labels to different stimulus images, which enables the network to infer the adversarial "directions" in the input space. 
In the experimental section, we demonstrate effectiveness of our approach on MNIST, CIFAR-10 and ImageNet pretrained models as targets. We use subset of TinyImageNet \cite{tiny120k} \cite{torralba200880} and Food-101 \cite{bossard14} as the stimulus for these targets.

\begin{figure}
\includegraphics[width=\columnwidth]{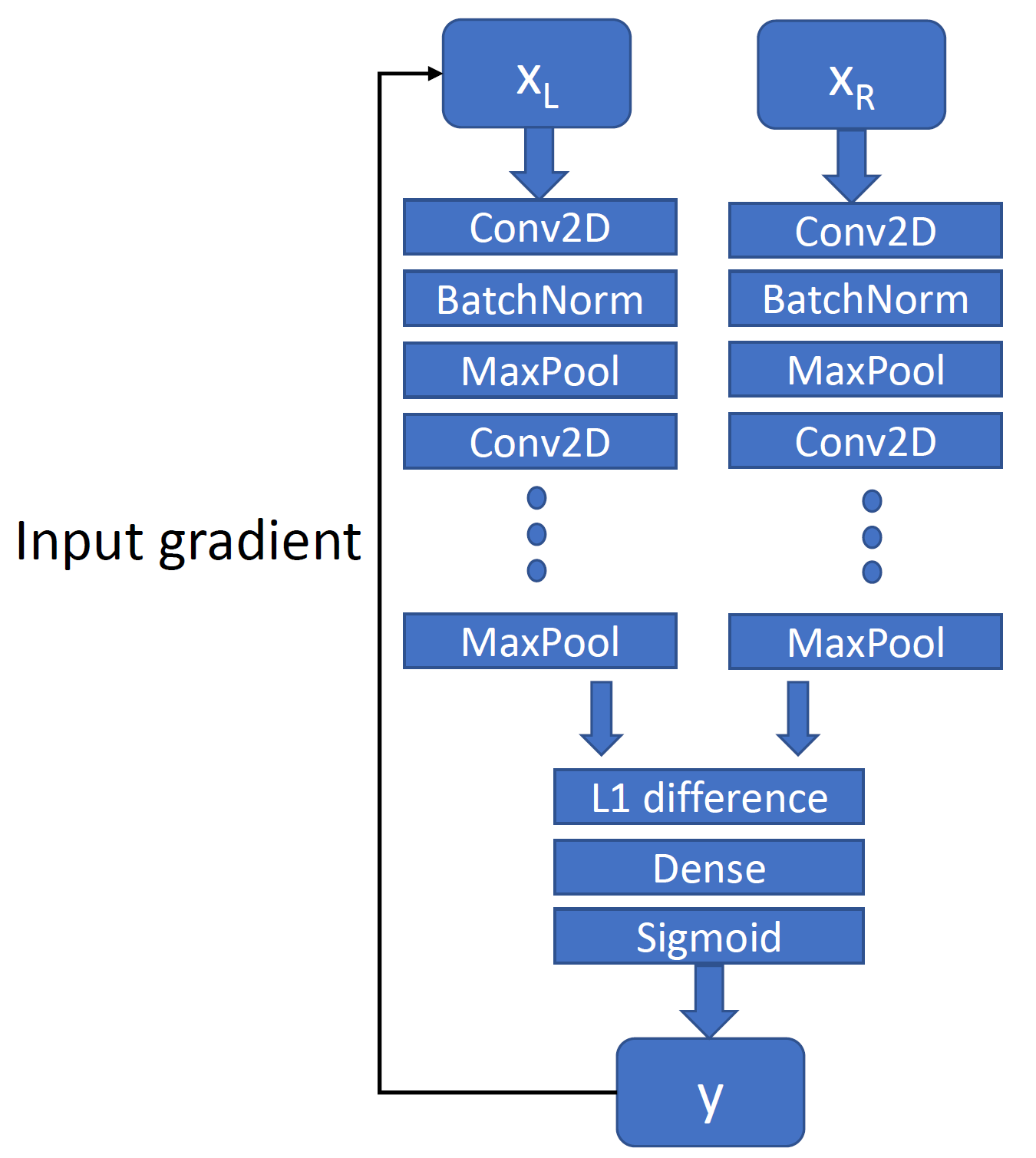}
\caption{\label{fig:inpt3} Siamese architecture for learning the adversarial perturbation.}
\end{figure}

\section{Proposed methodology}
This section describes our approach to generate adversarial examples for the target CNN.

\subsection{Training Siamese networks}
Siamese networks \cite{koch2015siamese}\cite{chopra2005learning} 
are well known architectures for learning data representations. 
 The training objective minimizes a loss function such that the features representing images from the same class have higher similarity whereas features for the images of different classes have low similarity. The mapping from the raw images to the feature space is facilitated by a convolutional network. They are proved to be 
efficient tools for learning the feature representation when the amount of labeled data is limited.

Since we do not assume knowledge of the target data distribution, we use an unlabeled mismatched dataset as the stimulus for the target model. During the experiments, for the MNIST target, we use TinyImageNet whereas for CIFAR-10 and ResNet targets, we use Food-101 as the stimulus. Initially, we get a small stimulus dataset labeled by the target CNN. 
To limit the number of queries, for each of our experiments, we only make 5k queries to the target CNN and collect the corresponding class labels. Since with $n$ image queries, we can generate $\frac{n^2}{2} - n$ unique image pairs, we can effectively train the Siamese network with a lower risk of overfitting.

Let $f_T$ denotes the function of the target CNN. Let $z \in [1,..,k]$ denote the label returned by the target model for the input $x$. Since the target is the classification model, $z$  is obtained as follows:
\begin{eqnarray}\label{eq:a1}
z_x = \arg \max f_T(x) 
\end{eqnarray}
It is feasible that the labels assigned to the stimulus may not cover all the possible classes that are present in the target training set. Since Siamese networks work on similarity/dissimilarity of the labels, this does not create an issue.
To avoid the possible class imbalance by random sampling, we generate an equal number of positive and negative pairs using the following procedure:

\begin{itemize}
\item Randomly pick a class from the stimulus labels. 
\item For that class, collect one positive and one negative example.
\item Iterate until the desired number of pairs are collected.
\end{itemize}


For training the Siamese network, we use a strategy similar to the one-shot learning approach proposed in \cite{koch2015siamese} where the L1 difference of the left and the right channel image representations is taken prior to the sigmoid unit.
Fig. \ref{fig:inpt3} shows an overview of the architecture of our Siamese network. 
Let $x_L$ and $x_R$ denote the input to the Left and to the Right subnetwork, respectively. Let $y$ denotes the ground truth label. If $x_L$ and $x_R$ belong to the same class, then $y = 1$, else $y = 0$. 
Let $f_S$ denotes the function represented by the Siamese. 
$f_S$ takes two images ($x_L$ and $x_R$) as the input and assigns a label to it as follows: 
\[
    y_{pred}= 
\begin{cases}
    1,& \text{if } f_S(x_L,x_R) >= 0.5 \\
    0,              & \text{otherwise}
\end{cases}
\]
Here, $y_{pred}$ denotes the label predicted for the given input image pair.
The network is trained with the binary cross entropy loss. Let $L^i$ denotes the loss for the $i^{th}$ sample 
\begin{eqnarray}\label{eq:a2}
L^i =  -y^i \log{y_{pred}^{i}} - (1 - y^i) \log{(1 - y_{pred}^{i})}
\end{eqnarray}
We use 10\% of the training pairs as the validation set and the training is terminated when the model's performance ceases to improve on the validation set. 

\subsection{Adversarial perturbations for Siamese}\label{sec:adv}
Our aim is to produce the minimally altered example of an image $x$, say $x^*$, such that the target model mis-classifies the example, i.e.
\begin{eqnarray}\label{eq:a1}
x^* = x + \delta \hspace{0.2cm} \text{s.t.} \hspace{0.2cm} \arg\max f_T(x) \neq \arg\max f_T(x^*)
\end{eqnarray}
However, we do not have access to $f_T$, and we now use the trained $f_S$ as the proxy.  We learn the \emph{adversarial perturbations} for $f_S$ and
show that, they are transferable w.r.t. the target model function $f_T$.



To generate adversarial examples for the Siamese network, we resort to the FSG method \cite{goodfellow2014explaining}. Since we have two images as input and a binary label as output, our attempt is to flip the label of the given input pair.
We take the gradient of $x_L$ w.r.t. the output of the network, i.e., $f_S(x_L,x_R)$. We iteratively modify $x_L$ so as to flip the initial estimated label. If initial value of $y_{pred}$ is 0, we need to perform the gradient ascent to flip it to 1. If initially, $y_{pred}$ is 1, we need to perform gradient descent to flip the label. 
To facilitate the switch between the ascent and the descent, we use the following update equation:
\begin{eqnarray}\label{eq:a2}
x_L = x_L + \epsilon_S \hspace{0.1cm} sign(0.5 - y_{pred}) \hspace{0.1cm} \frac{\partial f_S}{\partial x_L}
\end{eqnarray}
Here, $\epsilon_S$ denotes the weight assigned to the gradient.
Note that the term $sign(0.5 - y_{pred})$, automatically choses the ascent or descent appropriately.
As opposed to FSG for the multiclass case, which takes the gradient w.r.t. to the loss, we directly take the gradient w.r.t. the model output. 

The perturbation that resulted from the label flip is then obtained as
\begin{eqnarray}\label{eq:a2}
x_g = x_L - x^{init}_L
\end{eqnarray}
Here, $x^{init}_L$ denotes the initial value of $x_L$. 
The procedure is described in the Algorithm 1 block.

\begin{algorithm}
\caption{Learning adversarial perturbations using Siamese}
\begin{algorithmic}[1]
\State INPUTS: $x_L$, $x_R$, $y$, $\epsilon_S$, \text{max\textunderscore iter}
\State OUTPUT: adversarial perturbation: $x_g$

\State Initialization: $x_g = 0$, $x_L^{init} = x_L$ 

\State $y_{pred} = I(f_S(x_L,x_R) >= 0.5)$
\If {$y_{pred} = y$} 
\For{$\text{iter} = 1$ to $\text{max\textunderscore iter}$}
\State \hspace{0.4cm} $x_L$ = $x_L$ + $\epsilon_S \hspace{0.1cm} sign(0.5 - y_{pred}) \hspace{0.1cm} \frac{\partial f_S}{\partial x_L}$
\State \hspace{0.4cm} $y_{pred} = I(f_S >= 0.5)$
\State \hspace{0.4cm} if $y_{pred}$ $\neq$ $y$ :
\State \hspace{0.8cm} $x_g$ = $x_L$ - $x_L^{init}$ 
\State \hspace{0.8cm} break
\EndFor
\EndIf

\end{algorithmic}
For MNIST and CIFAR-10 experiments, we use $\epsilon_S$ = 0.001. For ImageNet, we use $\epsilon_S$ = 1. We use \text{max\textunderscore iter} = 100 for all experiments. 
\end{algorithm}




\begin{figure*} [!h]
\centering
\begin{tabular}{c c}

\includegraphics[width=150pt, height = 150pt]{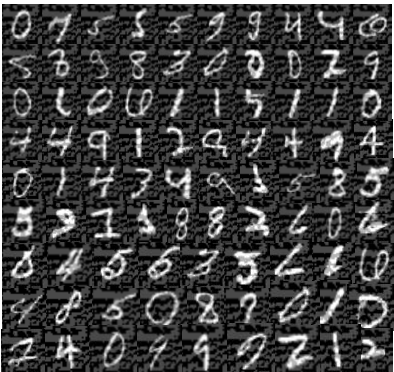}&
\includegraphics[width=\columnwidth]{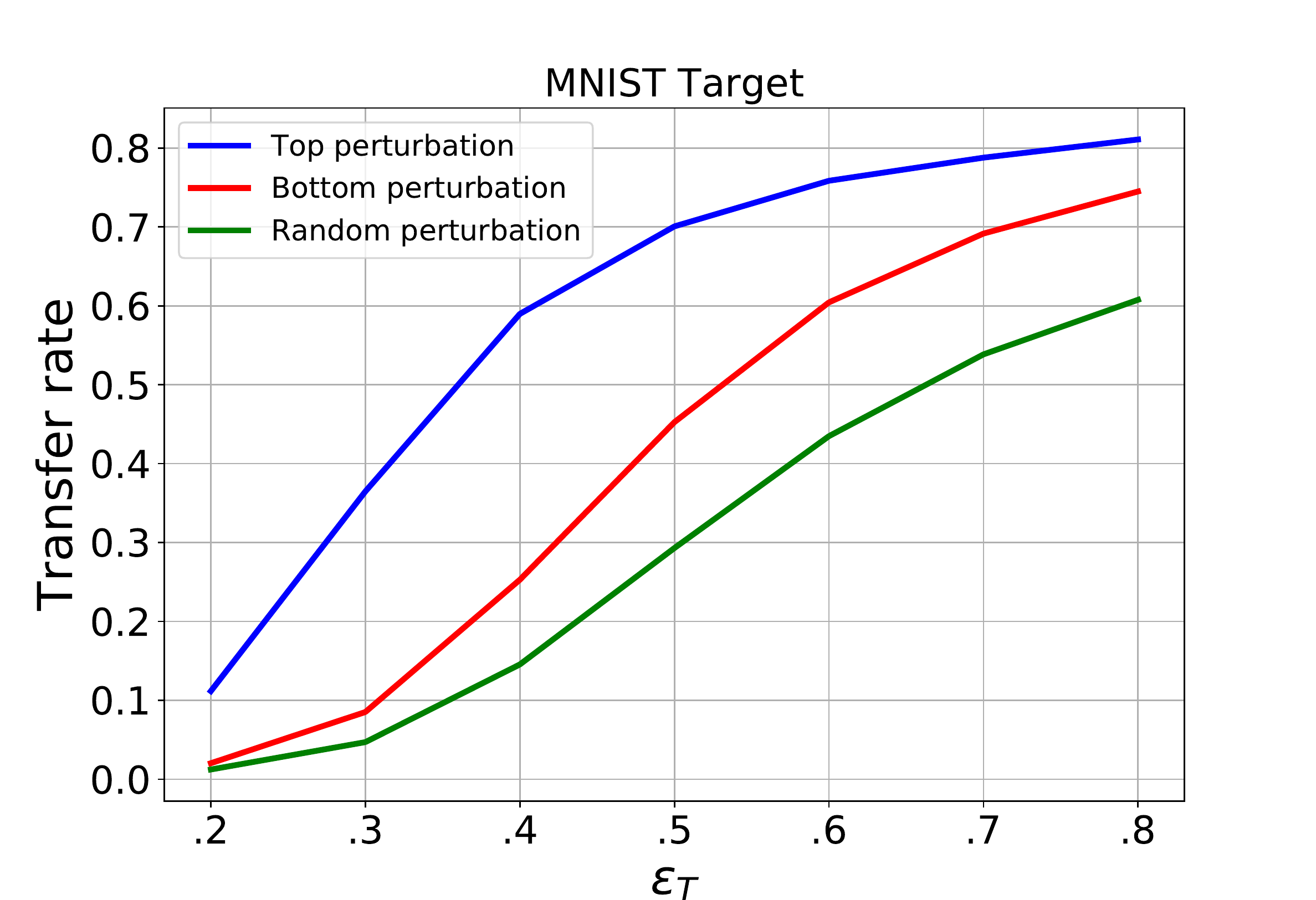}\\(a)&(b)\\

\includegraphics[width=150pt, height = 150pt]{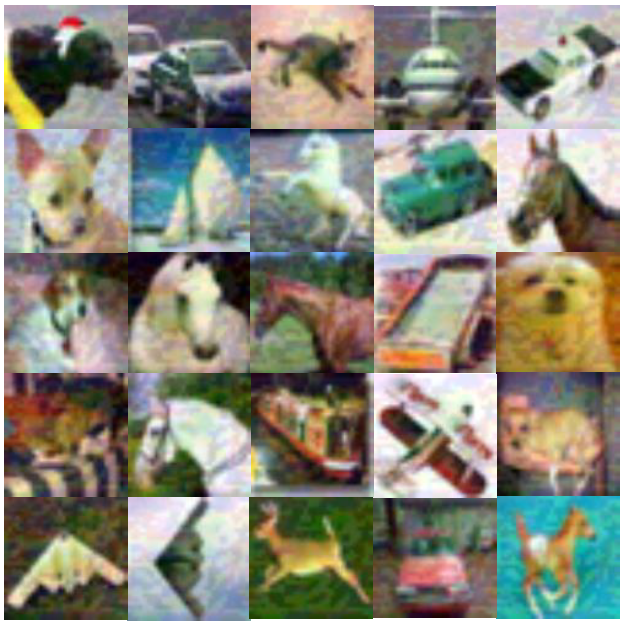}&
\includegraphics[width=\columnwidth]{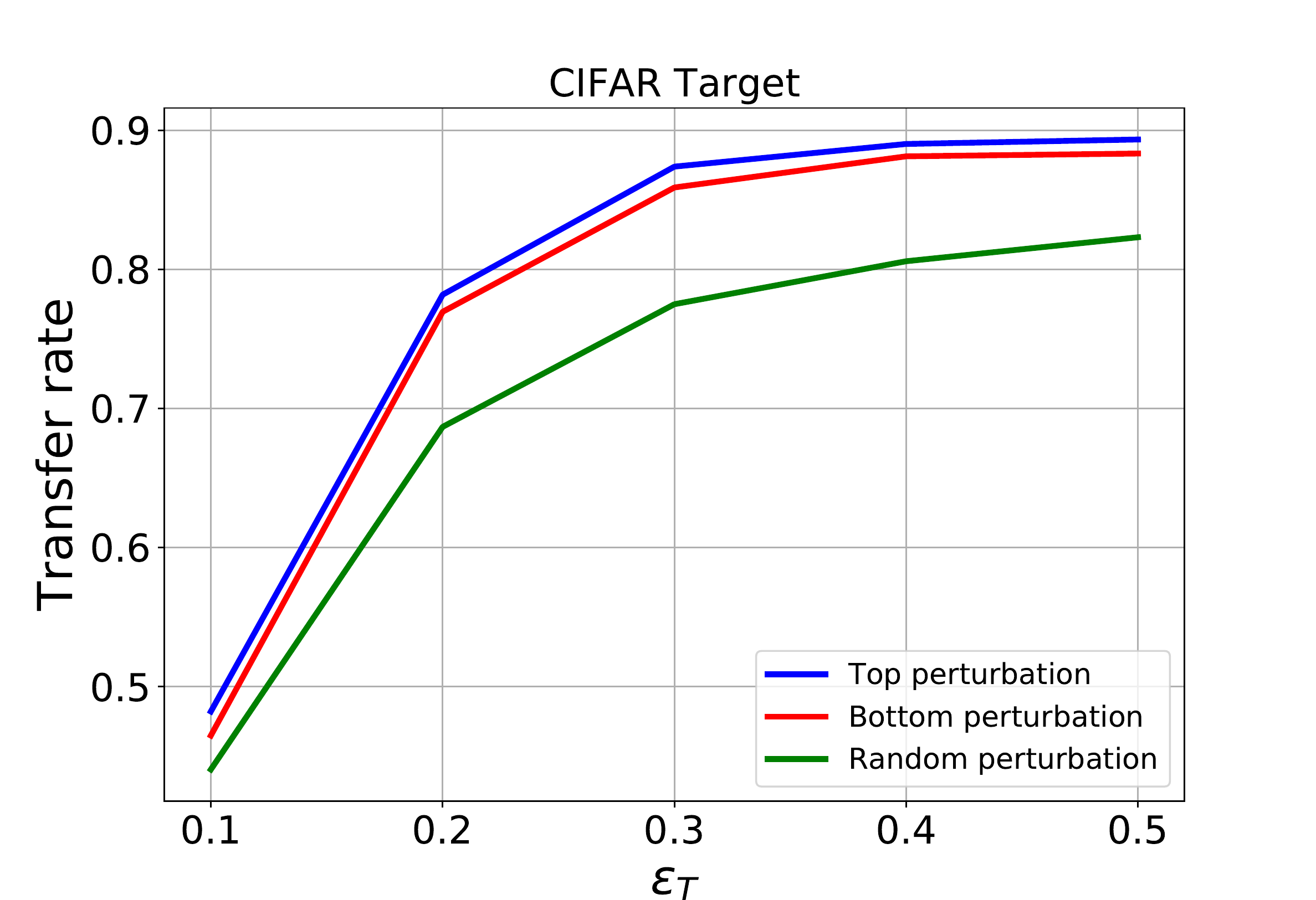}\\(c)&(d)\\

\includegraphics[width=\columnwidth]{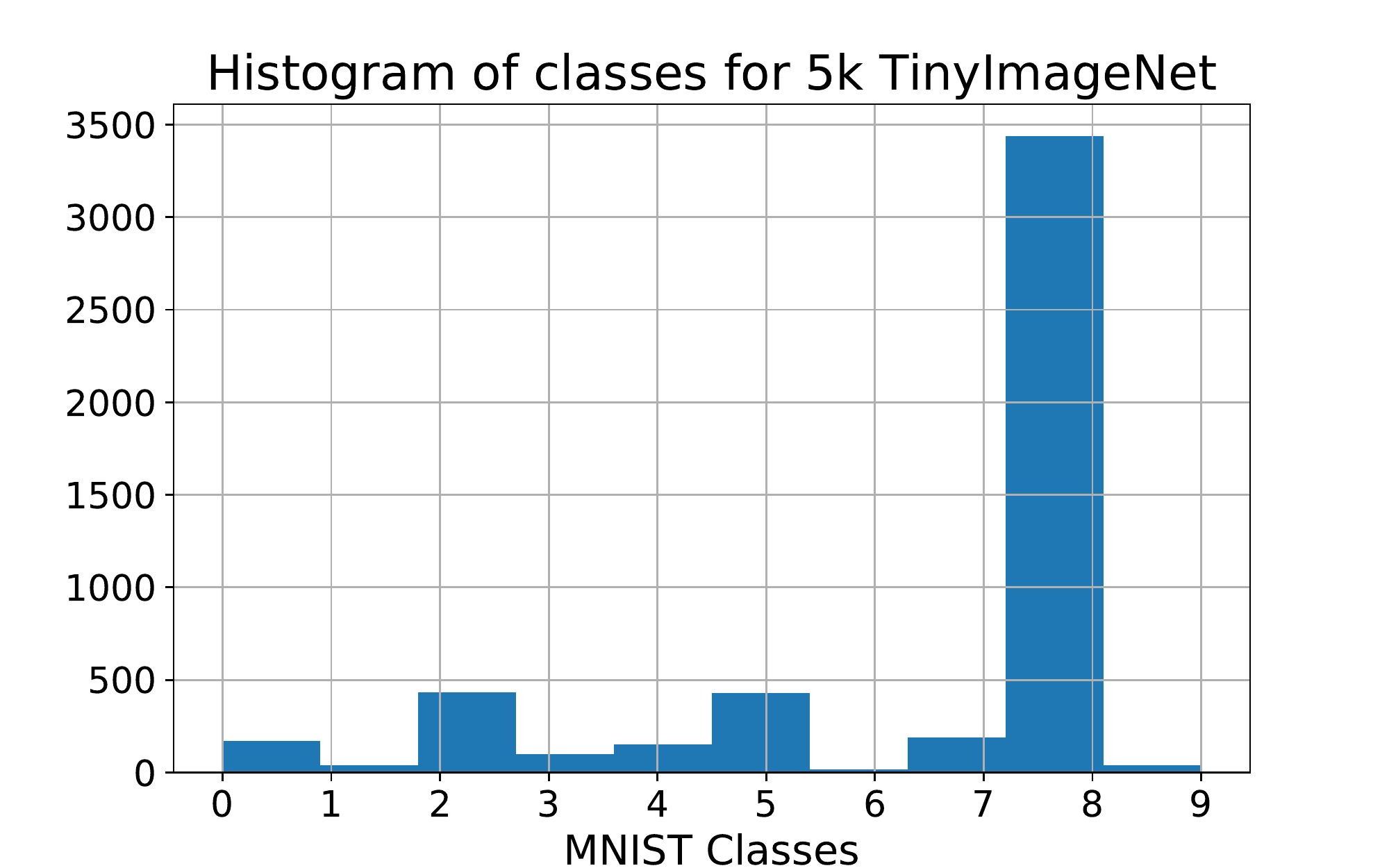}&
\includegraphics[width=\columnwidth]{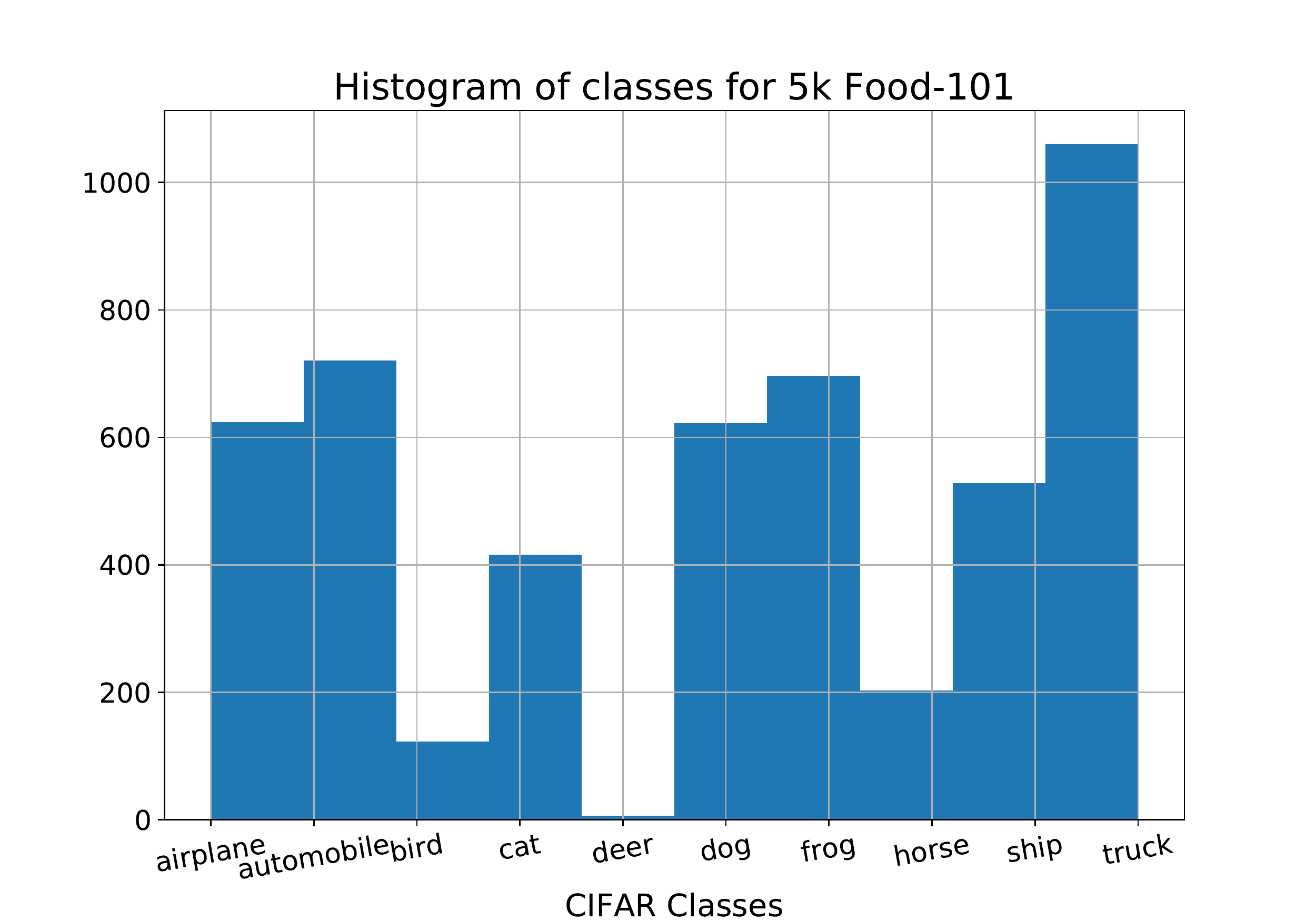}\\(e)&(f)\\

\end{tabular}
\caption{\label{fig:inpt0} Experiments with MNIST and CIFAR targets. (a) Adversarial images misclassified by the MNIST target for $\epsilon_T = 0.2$; (b) comparison of transfer rates for the top, the bottom and the random perturbation for MNIST target; (c) adversarial images misclassified by the CIFAR target for $\epsilon_T = 0.1$; (d) comparison of transfer rates for top, the bottom and the random perturbation for CIFAR target showing that the top ranked perturbations always have the higher transfer rate; (e) histogram of stimulus labels assigned by the MNIST target;(f) histogram of stimulus labels assigned by the CIFAR target.}
\end{figure*}

\subsection{Generating adversarial examples for target model}
We perturb target samples using $x_g$ as follows: 
\begin{eqnarray}\label{eq:add}
x_{P} = x_T + \epsilon_T \hspace{0.1cm} sign(x_{g})
\end{eqnarray}
Here $x_T$ denotes the sample from the target dataset and $x_P$ denotes the perturbed sample. $\epsilon_T$ denotes the scale factor for the signed perturbation.
We define the transfer rate $T_r$ as the percentage of labels changed by the learned perturbation:
\begin{eqnarray}\label{eq:a2}
T_r = \frac{\sum\limits_{i=1}^N{I(z^{i}_{x_{P}} \neq z^{i}_{x_{T}})}}{N} 
\end{eqnarray}
Here, $N$ denotes the size of the target set and $I$ denotes the indicator function whose output is 1 if the condition is true; else the output is 0.
Note that our aim is to \emph{change} the label of the target image that was previously assigned by the target model. The modified label can be any possible class. 
 
To more effectively learn the $x_g$, we generate a small set of test image pairs and obtain the perturbation for each by following the procedure described in Algorithm 1. 
We demonstrate that the perturbations that cause the maximum change in the loss (of the Siamese network) with the smallest infinity (max) norm, has the best transfer rate. 
We score each perturbation as follows: 
\begin{eqnarray}\label{eq:score}
s^{j} = (L^j_{x_L^{init} + x_g^j} - L^j_{x_L^{init}}) / ||x^j_g||_{\infty}
\end{eqnarray}
Here, $s^{j}$ denotes the score value assigned for the $j^{th}$ perturbation.
We rank perturbations in descending order of the the score value. We use the highest-ranked perturbation to generate the adversarial examples for the target set.
In the experimental section, we demonstrate the validity and the effectiveness of our scoring function.

\section{Experimental results}
We demonstrate effectiveness of our approach on pretrained target models trained on MNIST, CIFAR-10 and ImageNet datasets.


\begin{figure}
\includegraphics[width=\columnwidth]{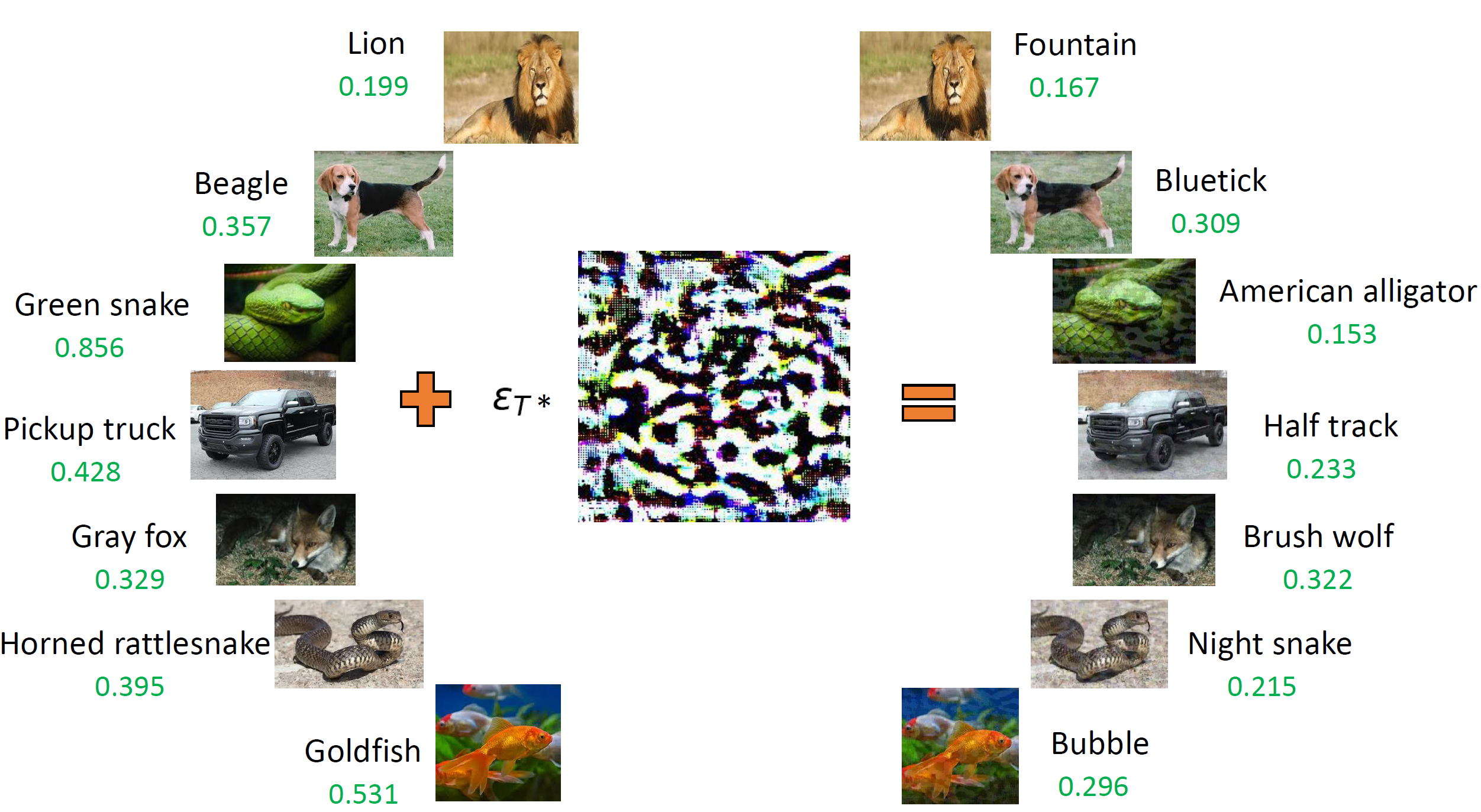}
\caption{\label{fig:inpt17} Adversarial images for ResNet-50. The center image shows the learned adversarial perturbation. Class probabilities are displayed in green. }
\end{figure}

\subsection{MNIST CNN target}
The MNIST dataset consists of gray-scale images of size $28 \times 28$ consisting of handwritten digits. We use a pretrained network as the MNIST target, which has approximately 184k parameters and test accuracy of 99.58\%.  
For MNIST target, we use 5k randomly chosen images from the TinyImageNet dataset \cite{tiny120k} \cite{torralba200880} as the stimulus. TinyImageNet is the natural image dataset and has different data distribution than the MNIST images. Stimulus images are resized and converted to gray-scale prior to querying. We obtain the multiclass labels on the stimulus using the MNIST target. Fig. \ref{fig:inpt0}(e) shows the histogram of labels assigned to the 5k stimulus dataset. Peculiarly, label 8 gets assigned the most. 

We train the Siamese network on 30k image pairs derived from these labels. 
From the training data, we use 10\% data as the validation set. The network architecture that performs well on the validation set is chosen.
The architecture of the subnetwork of the Siamese is \newline
[Conv(128,5,5)--BatchNorm--MaxPool(2)--Conv(128,5,5)--BatchNorm--MaxPool(2)--Conv(128,5,5)--BatchNorm--MaxPool(2)]. \newline
We use three convolution layers followed by batch-normalization and max-pooling layers. Each convolution layer has 128 filters of size $5 \times 5$. 
We use a dense layer of size 512 prior to the sigmoid unit.
Our network has approximately 1.4M parameters. We use batch-normalization layers and dropout to regularize the network. We observe that results are fairly robust against the small changes in the Siamese architecture.

Thereafter, we follow the procedure described in Algorithm 1 and learn the adversarial perturbations for a small set of newly generated image pairs. 
$\epsilon_S$ is set to 0.001 and $\text{max\textunderscore iter}$ is set to 100.
We rank the perturbations according to the scoring function defined in Eq. \ref{eq:score}.
We use the top-ranked perturbation as $x_g$ and perturb 10k images from MNIST test set according to Eq. \ref{eq:add}. 
Fig. \ref{fig:inpt0}(a) shows the adversarial images misclassified by the MNIST target. Note that we use the MNIST test set only to demonstrate the results for the adversarial examples. The MNIST data is not assumed during the training process.  

We now evaluate the effectiveness of our scoring strategy. We compare the transfer rates for the highest-(top-) ranked perturbation and the lowest- (bottom-) ranked perturbation. We also perform a comparison with a random perturbation to verify the superiority of learned perturbations. In case of random perturbation, we use a zero mean and unit standard deviation noise image as $x_g$. We obtain transfer rates with five random perturbations and plot the average transfer rate. Fig. \ref{fig:inpt0}(b) shows the plots of the transfer rates for different values of $\epsilon_T$. 
It can be seen that the top-ranked perturbation always performs better than the bottom ranked one. Also, learned perturbations have higher transfer rates than the random perturbation. 

Though, we get higher transfer rates with larger values of $\epsilon_T$, the perturbation becomes increasingly visible.
We get the transfer rate of 36.48\% for $\epsilon_T$ value of 0.3. For the similar value of $\epsilon_T$, Papernot et al. \cite{papernot2017practical} report the transfer rate of 78.72\% with the query dataset similar to the target dataset. Although our transfer rate is lower, it is interesting to note that it is obtained with a single perturbation applied to the entire test set with the smaller number of queries of mismatched dataset.


\subsection{CIFAR CNN target}
The CIFAR-10 dataset consists of color images of size $32 \times 32$.
For this experiment, we use a pretrained target network which has approximately 776k parameters and test accuracy of 82.16\%.  
For the CIFAR target, to have a mismatched set, we use 5k randomly chosen images from the Food-101 dataset \cite{bossard14} as the stimulus. 
The Food-101 dataset consists of images of food items corresponding to 101 different categories.
We obtain labels on the 5k stimulus set. Fig. \ref{fig:inpt0}(f) shows the histogram of stimulus labels assigned by the CIFAR target.
We generate 70k image pairs from the labeled stimulus to train the Siamese network. 
We use the same architecture for Siamese as for the MNIST experiment. The network has approximately 1.8M parameters.  
Fig. \ref{fig:inpt0}(c) shows the adversarial examples misclassified by the CIFAR target. 
Fig. \ref{fig:inpt0}(d) shows the comparison of transfer rates for the top, the bottom and the random perturbation. Here as well, for the random perturbation, we report average transfer rate over five different random perturbations. We observe the similar trend that the top-ranked perturbation performs the best. 

We now study the effect of the training set size for the Siamese network on the transfer rate. Table \ref{table:ensem} shows the transfer rates for the MNIST and CIFAR targets for varied number of input image pairs. We use five different subsets of test image pairs and calculate the mean and the standard deviation of the transfer rates. We observe that transfer rates increase with more number of pairs. However, we observed that after a certain number of pairs, transfer rates do not change significantly.

\begin{table}
  \small
  \begin{tabular}{llllssslsss}
    \toprule
    \multirow{2}{*}{\# image pairs} & 
              
      \multicolumn{1}{c}{MNIST transfer rate} & \multicolumn{1}{c}{CIFAR transfer rate}\\
                 
      &{(Tinyimagenet stim.)} & {(Food-101 stim.)} \\
      \midrule 
    10k & $0.435 \pm 0.074$  & $0.376 \pm 0.026$ \\
    20k & $0.511 \pm 0.055$  & $0.419 \pm 0.038$ \\
    30k & $0.635 \pm 0.059$  & $0.426 \pm 0.06$ \\

    \bottomrule
  \end{tabular}
  
  \caption{\label{table:ensem} Comparison of transfer rates with varying number of training image pairs. For MNIST target, $\epsilon_T$ is set to 0.5 whereas for CIFAR target, $\epsilon_T$ is set to 0.1.}
  
\end{table}

Experimental results demonstrate that our approach is able to learn the adversarial perturbations even under restricted setting. We observe that a single perturbation can be estimated that maximally perturbs the target set. This underlines the existence of universal perturbations, which we can effectively learn using Siamese networks. Next, we qualitatively demonstrate results of our approach on the pretrained target model trained on ImageNet dataset.

\subsection{ResNet target}
We use a pretrained ResNet-50 model as the target. For this target as well, we use 5k images from the Food-101 dataset as the mismatched stimulus. 
We generate 1M image pairs from the class labels obtained from the target and train the Siamese network. 

The architecture of the Siamese network is \newline
[Conv(128,5,5)--BatchNorm--MaxPool(2)--Conv(128,5,5)--BatchNorm--MaxPool(2)--Conv(128,5,5)--BatchNorm--MaxPool(2)--Conv(128,5,5)--BatchNorm--MaxPool(2)]. \newline

Note that, we use a similar architecture to those of the MNIST and CIFAR experiments where instead of three, we use four convolution/max pool layers. The network has approximately 14M parameters. 
We set $\epsilon_S$ = 1 and $\text{max\textunderscore iter}$ to 100.
Fig. \ref{fig:inpt17} shows the result of adding a learned adversarial perturbation to input images. The image in the center shows the highest-ranked perturbation. Images on the left show the original inputs and images on the right show the adversarial examples. $\epsilon_T$ is calculated separately for each image.  
Note that the same perturbation is able to switch the label of the inputs. For some images, the effect of the perturbation is visible in the output image, e.g., for the green snake image, $\epsilon_T$ had to be set to 27 to change the label. However, for most of the images, smaller values of $\epsilon_T$ effected the class change. In this case as well, we observed that our perturbation is superior to the random perturbation. With the random perturbation, a much larger value of $\epsilon_T$ was needed to effect the label change. Fig \ref{fig:inpt117} shows  another example of adversarial image generated by our approach for the image taken from  \cite{goodfellow2014explaining}.

\begin{figure}
\includegraphics[width=\columnwidth]{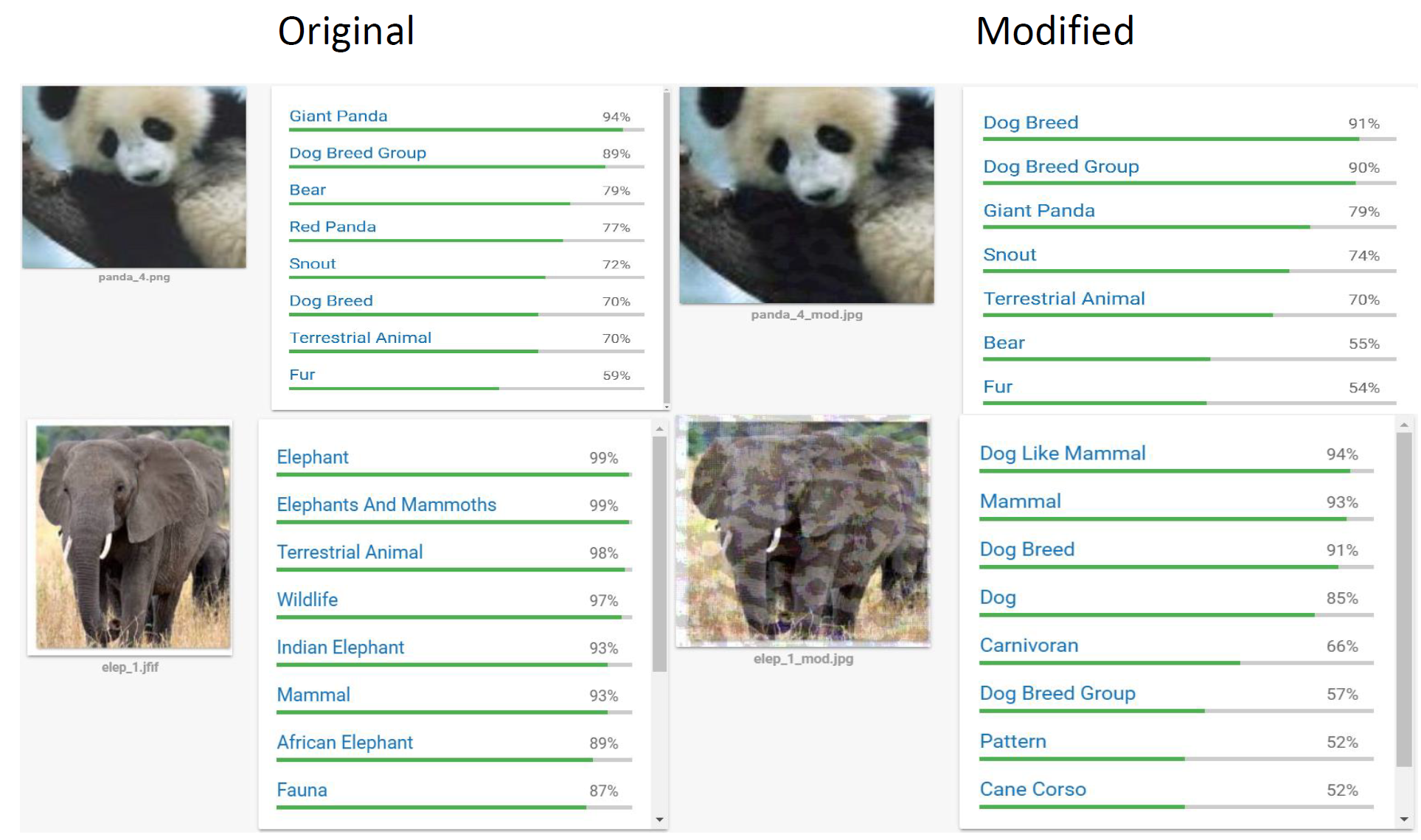}
\caption{\label{fig:inpt116} Experiment with the Google vision API. Left and right columns show output of the vision API for original and modified images, respectively. }
\end{figure}

Since adversarial images are known to be transferable between the models, we performed an experiment with the Google vision API. Fig. \ref{fig:inpt116} shows the class labels returned by the vision API for the original images and the perturbed images generated by our approach. Perturbed images are misclassified by the vision API. The result indicates that the adversarial perturbation directions computed by our approach are effective.

\section{Conclusion}
In this paper, we proposed an approach based on Siamese networks to generate adversarial examples under a black-box setting. Since Siamese networks are trained on image pairs, with the limited number of mismatched queries to the target model, we could effectively learn the adversarial directions in the input space. We observed that the perturbations that maximally alter the loss with minimum max norm are the most effective. 
Experiments with MNIST, CIFAR-10 and ImageNet targets with TinyImageNet and Food-101 stimulus demonstrated the efficacy of the proposed approach.

\bibliographystyle{IEEEtran}
\bibliography{adversial}

\end{document}